\documentclass[10pt,twocolumn,letterpaper]{article}

\usepackage[pagenumbers]{cvpr}

\usepackage[T1]{fontenc}
\usepackage[utf8]{inputenc}
\usepackage{amsmath,amssymb,booktabs,graphicx,tabularx,array,multirow}
\usepackage{cuted}
\usepackage{xcolor}
\usepackage{url}

\newcommand{\model}{DuoFlow}
\newcommand{\beatc}{MotionSpeech}

\usepackage[pagebackref,breaklinks,colorlinks,allcolors=blue]{hyperref}

\title{DuoFlow: Towards Synchronized Speech and Motion Generation with Fine-Grained Control}

\author{%
Jun Dong\textsuperscript{1} \quad Ronghui Li\textsuperscript{1} \quad Zhongyuan Hu\textsuperscript{1} \\
Zhao Yunqi\textsuperscript{1} \quad Lijie Guo\textsuperscript{1} \quad Ruizhang Xu\textsuperscript{1} \\
Di Kang\textsuperscript{2} \quad Linchao Bao\textsuperscript{2} \quad Xiu Li\textsuperscript{1} \\
\textsuperscript{1}Tsinghua University \quad \textsuperscript{2}Tencent%
}

\begin{document}
\maketitle
\begin{strip}
\centering
\includegraphics[width=0.84\textwidth]{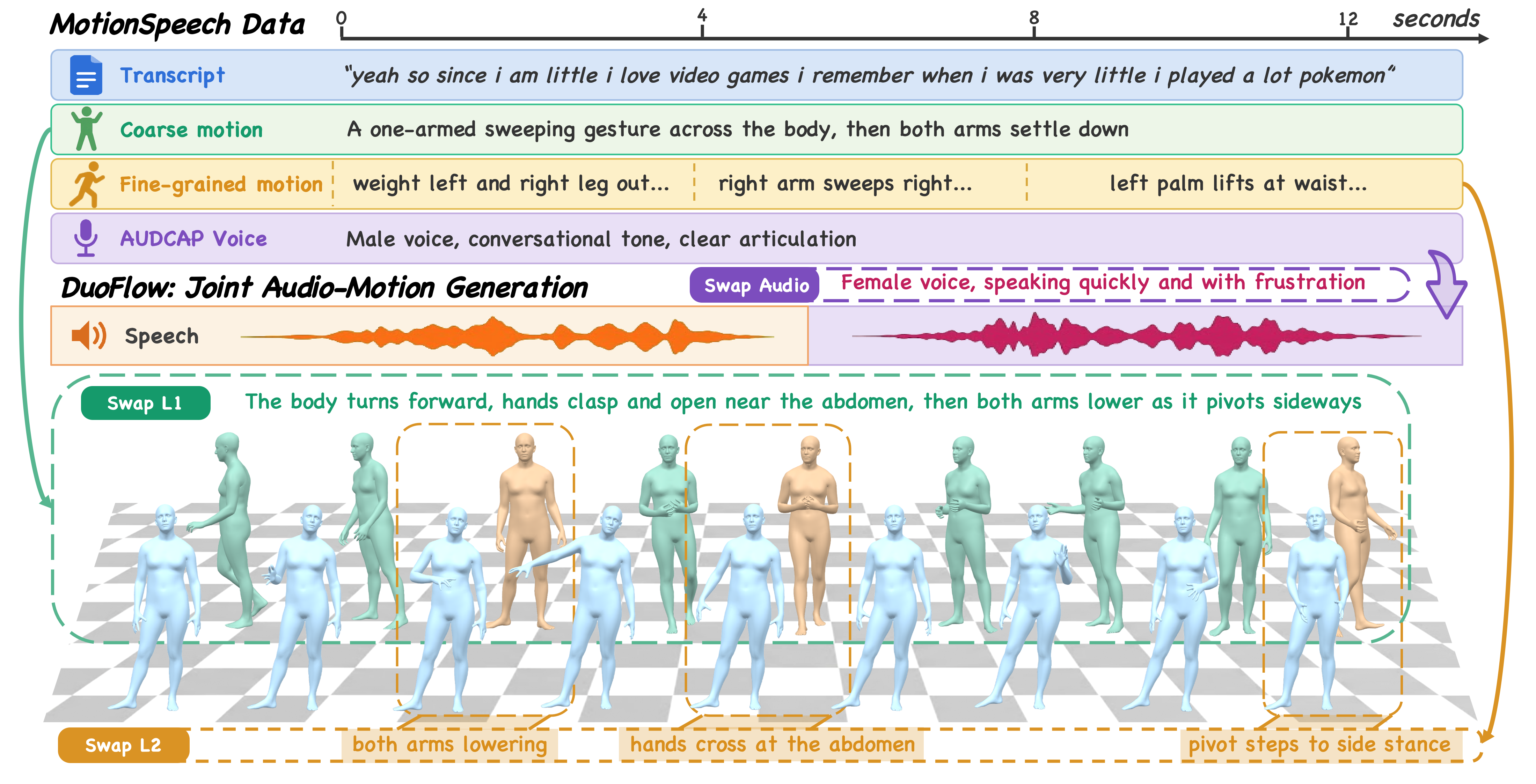}
\captionsetup{hypcap=false}
\captionof{figure}{Our \textbf{MotionSpeech} dataset provides approximately 125 hours of synchronized speech and full-body motion, annotated with coarse- and fine-grained motion descriptions, transcripts, and comprehensive voice descriptions. \textbf{\model{}} enables controllable joint speech--motion generation from text conditions. Modifying any modality description can guide generation: speech content (blue), coarse overall motion intent (green), fine-grained temporal motion (yellow), and vocal expression style (purple).}
\captionsetup{hypcap=true}
\label{fig:teaser}
\end{strip}

\begin{abstract}
An embodied speaker's expressiveness depends not only on what is said, but also on how the voice sounds and how the body moves. However, existing joint speech--motion generation models are largely driven implicitly by transcript semantics or intermediate speech representations, offering little explicit control over motion hierarchy, body parts, or vocal style. To bridge this gap, we construct \beatc{}, a structured benchmark for controllable joint speech--motion generation that includes transcripts, voice-style descriptions, and motion conditions at two hierarchical levels. Building on this benchmark, we propose \model{}, a dual-stream latent flow transformer that jointly denoises speech and motion under a shared noise level and a unified second-based temporal coordinate. Modality-specific residual connectors inject structured conditions into the corresponding streams, layer-wise bidirectional cross-modal attention coordinates their generation, and a monotonic phoneme pathway combines normalized progress anchors with CTC supervision to strengthen speech temporal modeling. \model{} provides a unified interface for fine-grained motion and vocal-style control and demonstrates the ability to synthesize high-quality, synchronized, and instruction-consistent speech--motion sequences.~\href{https://jundong-dev.github.io/DuoFlow-project-page/}{Project page}.
\end{abstract}

\section{Introduction}
\label{sec:intro}

An embodied speaker communicates through speech and movement at the same time: an utterance may be accompanied by a rhythmic beat, an illustrative hand gesture, or a subtle postural shift~\citep{nyatsanga2023review,liu2022beat}. In many applications the verbal content is fixed in advance, as in a script, a voice-over, or an avatar's dialogue. The central question then shifts from what is said to how it is delivered: how the utterance should sound, and how the body should move as it is spoken. Because conversational gesture varies considerably across speakers and contexts~\citep{ginosar2019learning,yoon2020speech}, a single plausible realization is insufficient when a particular delivery is intended, and user intent can be difficult to reconcile with fine-grained motion and timbre. In practice, users can readily express an overall gestural intent but cannot be expected to enumerate the temporally ordered actions of individual body parts. A useful control interface should therefore cover hierarchical levels of granularity, expanding concise high-level guidance into detailed, coherent motion while also following fine-grained instructions when they are available.

Existing systems generally lack such an interface because control remains fragmented across two largely separate paradigms. On the speech side, instruction-following text-to-speech can follow vocal instructions but does not synthesize body motion~\citep{du2024cosyvoice2}. On the motion side, prompt-based full-body control, caption-guided gestures, and joint-level spatial constraints have substantially advanced motion controllability~\citep{chen2024syntalker,fang2025coordspeaker,pinyoanuntapong2026interactgesture}, but still use speech that has already been produced as the conditioning input. Cascading a speech generator with a motion generator combines the two outputs but retains a one-way interface: the motion stage receives completed audio, so a motion instruction supplied at that stage cannot shape the speech upstream. Early and recent joint speech--gesture studies have demonstrated the feasibility of joint generation~\citep{alexanderson2021spontaneous,wang2021integrated,mehta2024match,guo2024fasttalker,guichoux2026gelina,ma2026motionomni}; joint speech--face generation~\citep{kwon2026jamflow,lim2026faceplex} and dialogue systems supporting discrete body-action directives~\citep{zhang2026vibes} provide related precedents. However, these interfaces still do not allow users to control both motion and vocal expression through instructions.

Turning joint generation into controllable generation raises three challenges. First, supervision for such an interface does not yet exist. Existing co-speech corpora provide audio, transcripts, motion, and in some cases semantic or emotion labels~\citep{liu2022beat,liu2024emage}, while caption-guided motion work adds textual gesture supervision~\citep{fang2025coordspeaker}; however, none aligns the same speech--motion target with hierarchical motion descriptions and a voice description. Second, semantic condition guidance is complex: speech and motion differ in temporal resolution, so their conditions must be injected accurately, and the model must bridge the gap between coarse user intent and fine-grained motion control. Third, evaluation remains incomplete: existing work~\citep{kucherenko2023genea,liu2024emage,nagy2026genea} does not directly evaluate whether jointly generated speech and motion follow their respective voice and motion instructions, motivating a unified benchmark for quality evaluation.

To this end, we construct \beatc{} and propose \model{} (Figure~\ref{fig:teaser}). \beatc{} builds on existing public datasets~\citep{liu2024emage,agrawal2025seamless} and refines four high-quality annotation channels: a coarse motion description (L1), a fine-grained description of body-part actions and their temporal order (L2), a voice description (AUDCAP), and a transcript. On this basis, we propose \model{}, a dual-stream latent flow transformer. The model denoises motion and audio latents in parallel under a shared noise level and a unified time coordinate measured in seconds. Bidirectional cross-modal attention in every block allows the two streams to exchange information, while modality-specific residual connectors adapt the structured condition for each stream. A dedicated monotonic phoneme-alignment pathway adds a normalized progress anchor to the phoneme context and uses phoneme-CTC supervision to make intermediate audio states follow the transcript order without forced phoneme boundaries. Progressive audio-only pretraining followed by joint speech--motion training gives the model strong generalization capability.

Our contributions are threefold. (i) We present \beatc{}, a four-channel structured benchmark for fine-grained controllable joint generation, comprising 34{,}950 clips (approximately 125.2 hours). (ii) We propose \model{}, the first joint speech--motion generation framework to support hierarchical motion and voice-condition control. (iii) We establish a unified evaluation protocol for generation quality, cross-modal synchronization, and text--motion consistency. Experiments show that \model{} is the only one with a native unified interface that simultaneously supports end-to-end joint generation, fine-grained motion control, and free-form vocal-style control, while maintaining strong motion quality, speech--motion coordination, and instruction consistency.

\section{Related Work}
\label{sec:related}

\subsection{Co-speech motion generation.}
Given speech, co-speech models synthesize body movement: early cross-modal translation~\citep{ginosar2019learning}, trimodal text--audio--speaker conditioning with the FGD metric~\citep{yoon2020speech}, and large annotated corpora with semantic baselines~\citep{liu2022beat,liu2024emage}. Architectures advanced from regression to diffusion, state-space, and flow models~\citep{liu2024emage,xu2024mambatalk,liu2025gesturelsm}. Recent systems add caption- or example-level semantics~\citep{fang2025coordspeaker,zhao2026personagest,rosenthal2026signature,liu2026semconflow}. However, these methods all treat speech as a conditioning input and do not support joint speech--motion generation.

\subsection{Joint speech--motion generation.}
Match-TTSG~\citep{mehta2024match} pioneered joint flow synthesis of speech and upper-body gesture in one process; FastTalker~\citep{guo2024fasttalker} reaches real-time joint generation by reusing speech features; Gelina~\citep{guichoux2026gelina} unifies the two modalities with interleaved discrete tokens and style cloning; Motion-Omni~\citep{ma2026motionomni} emits full-body motion natively from dialogue hidden states with an open evaluation protocol; RapVerse~\citep{chen2025rapverse} targets vocals and whole-body dance; ViBES~\citep{zhang2026vibes} lets a dialogue agent accept discrete body-action directives; DyaPlex~\citep{nagano2026dyaplex} targets streaming duplex conversation. Adjacent joint models cover faces, audio-video, or agent stacks~\citep{chen2024diffsheg,jang2024faces,kwon2026jamflow,cheng2026unison,deng2026umind,zhang2026exomni,jiang2024solami}. Connectionist temporal classification (CTC) marginalizes monotonic paths between frame sequences and unsegmented label sequences~\citep{graves2006ctc}; we use it only as an auxiliary training signal on latent audio states, rather than as an autoregressive decoder or an inference-time forced aligner. These systems establish that joint generation is possible and increasingly efficient, but their interfaces either expose no user-facing motion semantics at all (Match-TTSG, FastTalker, Gelina, Motion-Omni) or trigger discrete behaviors instead of span- and part-level instructions (ViBES).

\subsection{Controllable multimodal generation.}
On the speech side, CosyVoice~2~\citep{du2024cosyvoice2} provides streaming zero-shot cloning and instruction-following speech, and MGM-Omni~\citep{wang2025mgmomni} stabilizes long-horizon personalized timbre; both control a single modality. LTX-2~\citep{hacohen2026ltx2} generates synchronized audio-video with a two-stream DiT and modality-CFG, and unified music--motion models~\citep{yang2025unimumo,you2024momu} align rhythmic latents across modalities; pixel-space audio-video and music-driven synchronization, however, do not provide semantically structured 3D motion amenable to part- and span-level instructions. Streaming systems optimize latency rather than controllability~\citep{zhu2026omniliveavatar,ji2026discoforcing}. \model{} follows this line of research by adopting joint latent flow generation as its mechanism and constructing a unified control interface.

\section{The \beatc{} Benchmark}
\label{sec:beatc}

\subsection{Data and Hierarchical Annotations}
\beatc{} couples full-body conversational motion with multi-scale hierarchical spoken-language annotations.
It selects and integrates samples from two complementary sources of natural talking--gesturing behavior:
(i) \textbf{BEAT2}~\citep{liu2024emage}, a studio-capture corpus pairing SMPL-X body motion with expressive English speech, and
(ii) \textbf{Seamless Interaction}~\citep{agrawal2025seamless}, an improvised conversational corpus of spontaneous multi-speaker dialogues with rich prosodic variation.

\begin{figure}[t]
\centering
\includegraphics[width=\columnwidth]{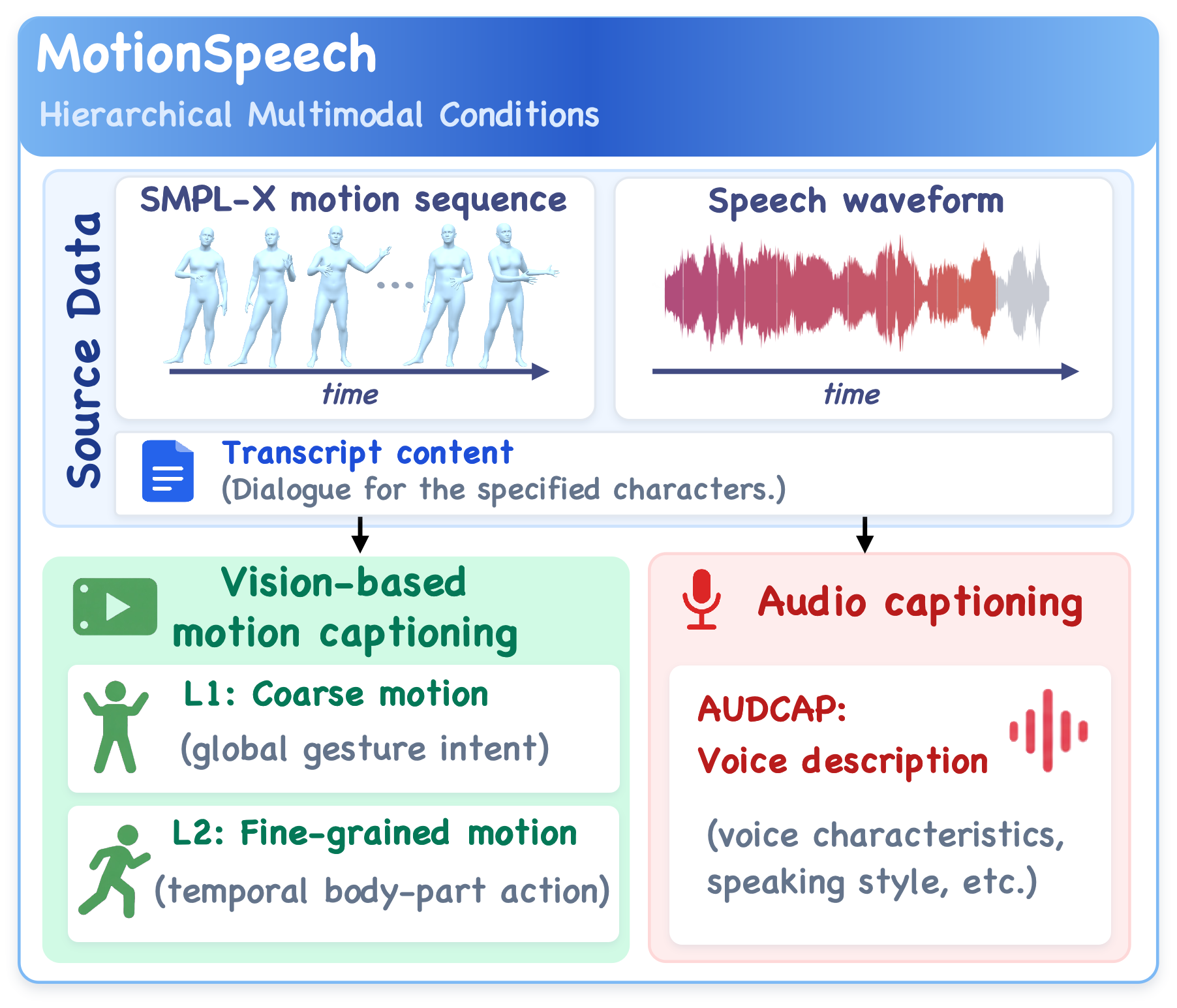}
\caption{\textbf{MotionSpeech data and annotation construction.} Each source segment pairs an SMPL-X motion sequence with its speech waveform and transcript. The annotation pipeline adds hierarchical motion descriptions and a voice description, followed by manual verification and correction.}
\label{fig:data}
\end{figure}

Each segment is annotated with a four-channel structured caption: (1) \textbf{L1 coarse motion}, describing overall body dynamics and gestures; (2) \textbf{L2 fine-grained motion}, describing temporally ordered body-part actions and directions; (3) the \textbf{re-recognized transcript}; and (4) \textbf{AUDCAP voice}, describing timbre and delivery style. The four tagged channels can be parsed independently.

\subsection{Dataset Construction}
\beatc{} is constructed in three steps.

\noindent\textbf{(i) Segmentation and pairing.}
Both source corpora are cut into uniform 387-frame windows with paired SMPL-X motion sequences and waveforms.
Motion is represented as $387\times333$ features (global translation plus 55 joints in 6D rotation), and audio is resampled to 16\,kHz mono.

\par\noindent\textbf{(ii) Hierarchical caption annotation.}
For every segment, GLM-5.3-Flash watches the rendered motion and writes (a) an \textbf{L1} summary of coarse posture and gesture style and (b) an \textbf{L2} sequential, body-part-level description. Qwen3-Omni-30B-A3B-Instruct analyzes the speech waveform to produce the \textbf{AUDCAP} voice description, while the spoken content is retained as the \textbf{re-recognized transcript} channel. Human annotators then inspect every four-channel record against the source motion and audio and correct inaccurate, incomplete, or malformed descriptions.

\par\noindent\textbf{(iii) Tokenization and latent precomputation.}
Before training, phoneme sequences are produced by a grapheme-to-phoneme tokenizer; motion and speech are encoded into latents by the motion VAE (Section~\ref{sec:motion_vae}) and the AudioLDM2 VAE~\citep{liu2024audioldm2}, respectively; text conditions are encoded by the Gemma text encoder and mapped by a feature-fusion module into trainable joint-condition features.
\par

\subsection{Statistics}
Table~\ref{tab:dataset_comparison} compares \beatc{} with representative co-speech motion datasets spanning studio motion capture, in-the-wild video reconstruction, group interaction, and situated conversation. Among the listed corpora, \beatc{} is the only one with paired natural-language motion captions and voice captions for every training window; Seamless Interaction provides visual-behavior descriptions for only a subset and does not annotate vocal timbre or delivery in natural language.
The released benchmark contains \textbf{34{,}950 segments} ($\approx$\textbf{125.2} hours), including \textbf{17{,}907 BEAT2} clips and \textbf{17{,}043 Seamless Interaction} clips.

\begin{table}[t]
\centering
\caption{\textbf{Public co-speech motion dataset comparison.} Mot. cap. and Voice cap. denote natural-language motion and voice descriptions, respectively. Partial specifically refers to Seamless's 3P-V visual-behavior descriptions released for an annotated subset; -- indicates unavailable data.}
\label{tab:dataset_comparison}
\scriptsize
\setlength{\tabcolsep}{1.2pt}
\renewcommand{\arraystretch}{0.96}
\begin{tabularx}{\columnwidth}{@{}>{\raggedright\arraybackslash}X>{\raggedright\arraybackslash}Xrrccc@{}}
\toprule
Dataset & Motion representation & Hrs. & Spk. & Tr. & Mot. & Voice \\
\midrule
Trinity~\citep{ferstl2018recurrent} & 3D body + hands & 4.00 & 1 & \checkmark & -- & -- \\
Talking With Hands~\citep{lee2019talking} & 3D body + hands & 50.00 & 50 & -- & -- & -- \\
TED Gesture~\citep{yoon2020speech} & 3D upper body & 97.00 & 1{,}766 & \checkmark & -- & -- \\
PATS~\citep{ahuja2020style} & 2D keypoints & 251.00 & 25 & \checkmark & -- & -- \\
BEAT~\citep{liu2022beat} & BVH + ARKit & 76.00 & 30 & \checkmark & -- & -- \\
SHOW~\citep{yi2023talkshow} & SMPL-X + face & 26.90 & 4 & -- & -- & -- \\
BEAT2~\citep{liu2024emage} & SMPL-X + FLAME & 60.00 & 30 & \checkmark & -- & -- \\
Seamless~\citep{agrawal2025seamless} & SMPL-H + face & 4{,}065.00 & 4{,}284 & \checkmark & Partial & -- \\
TED4C-L~\citep{gjaci2026sicage} & 3D upper body (9 joints) & 106.45 & 764 & \checkmark & -- & -- \\
\midrule
\beatc{} (ours) & SMPL-X (55 joints) & 125.20 & 30+ & \checkmark & \checkmark & \checkmark \\
\bottomrule
\end{tabularx}
\end{table}

\section{DuoFlow}
\label{sec:method}

\begin{figure*}[t]
\centering
\includegraphics[width=\textwidth]{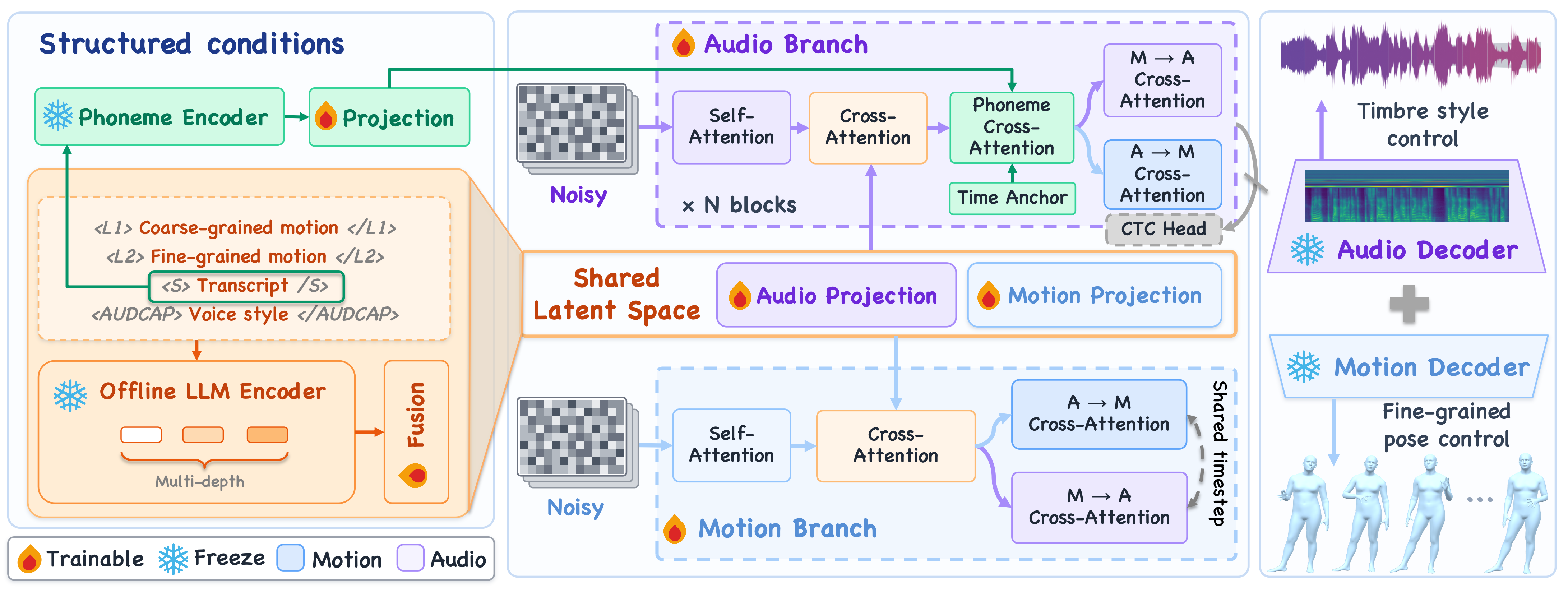}
\caption{\textbf{Overview of \model{}.} The selected motion instruction, transcript, and voice description form a shared semantic condition for two coupled latent streams. A frozen phoneme encoder supplies a separate audio condition. Its normalized phoneme-progress anchor preserves token order, and phoneme-CTC heads supervise the audio states at multiple depths during training only. Each dual-stream block combines self-attention, condition cross-attention, phoneme cross-attention, and bidirectional speech--motion exchange. Frozen decoders reconstruct the denoised latents as a waveform and a motion sequence.}
\label{fig:arch}
\end{figure*}

\subsection{Task and Model Overview}
For each segment, \beatc{} provides a transcript $s$, a voice description $c^{\mathrm{voice}}$, and paired motion descriptions $c^{\mathrm{L1}}$ and $c^{\mathrm{L2}}$. The model receives exactly one motion description at a time. Letting $\ell\in\{\mathrm{L1},\mathrm{L2}\}$ denote the selected granularity, its structured condition is $c_\ell=\big(c^\ell,s,c^{\mathrm{voice}}\big)$.
Conditioned on $c_\ell$, \model{} learns the joint distribution of a speech waveform $y^a$ and a synchronized full-body motion sequence $y^m$. Frozen modality encoders map paired speech and motion to continuous latents $z_0^a$ and $z_0^m$. A coupled transformer estimates both latent trajectories in one denoising process, after which the corresponding frozen decoders recover the waveform and motion. Keeping two streams allows each modality to retain its own representation and temporal resolution, while coupling them throughout generation coordinates delivery decisions before either output is finalized.

\subsection{Motion VAE}
\label{sec:motion_vae}
We adapt the MotionStreamer causal codec~\citep{xiao2025motionstreamer} to encode full-body motion into continuous latents. Each 30-Hz SMPL-X frame contains root translation and 55 joint rotations in 6D, yielding 333 features. After normalization, independent body, left-hand, and right-hand VAE branches encode the corresponding channel groups into 128, 64, and 64 latent channels. Causal convolutions compress time by a factor of three; concatenating the branches gives a 256-dimensional, 10-Hz motion sequence. The codec is trained with calibrated reconstruction and KL losses, with extra emphasis on root translation and rotation. We freeze it during flow training, use posterior samples as motion targets, and decode the generated latents back into SMPL-X motion.

\subsection{Structured Input Conditions}
The shared semantic condition $c_\ell$ combines the selected motion description, transcript, and voice description (AUDCAP). The transcript also provides a separate phoneme condition for the audio stream. Together, these conditions specify what to say, how the voice should sound, and how the body should move.

\noindent\textbf{Hierarchical motion conditions.}
Each annotated segment $i$ is expanded into two training instances, $\big(c_{i,\mathrm{L1}},y_i^m,y_i^a\big)$ and $\big(c_{i,\mathrm{L2}},y_i^m,y_i^a\big)$. The L1 instance asks the model to recover detailed motion from a concise global intent, whereas the L2 instance exposes temporally ordered body-part actions as a direct fine-grained control interface. Both instances share the same speech--motion target.

\noindent\textbf{Textual condition composition.}
For either motion level, we serialize the selected L1 or L2 motion description, followed by the transcript and AUDCAP. A frozen text encoder~\citep{gemmateam2025gemma3} extracts three complementary representations: the token-embedding output $H^{\mathrm{emb}}$, the first Transformer-layer output $H^{\mathrm{first}}$, and the final-layer output $H^{\mathrm{last}}$. These representations preserve the level distinction and lexical details while providing contextualized semantics, and are fused by a learned projection as $H=P\!\left([H^{\mathrm{emb}};H^{\mathrm{first}};H^{\mathrm{last}}]\right)$.

\noindent\textbf{Modality-specific condition adapters.}
The shared representation is adapted independently for motion and audio. For $r\in\{m,a\}$, a modality-specific connector produces $H^r=\operatorname{LN}\!\left(H+C_r(H)\right)$. The residual path retains token-level distinctions from the text encoder, while $C_r$ learns how the same motion, transcript, and voice conditions should be interpreted by each stream. Both streams attend to their adapted semantic condition throughout the transformer.

\noindent\textbf{Phoneme conditioning and progress.}
To efficiently preserve pronunciation, we convert transcript $s$ into $N$ valid phoneme IDs, encode them with a frozen VITS text encoder~\citep{kim2021vits}, and project the features to the audio-stream width. Because the encoder has no utterance-level position signal, we add a normalized progress anchor to strengthen order awareness in the cross-attention context. For valid phoneme $j$, let $p_j=(j+1)/N$ and define
\begin{equation}
\begin{aligned}
\gamma(p_j)&=\big[\sin\theta_{jk};\cos\theta_{jk}\big]_{k=1}^{32},\\
\theta_{jk}&=2\pi p_j\,2^{9(k-1)/31}.
\end{aligned}
\label{eq:phoneme_anchor}
\end{equation}
A learned projection $A$ maps this 64-dimensional code to the audio width, giving the anchored phoneme context $\widetilde e_j^p=P_p e_j^p+A\gamma(p_j)$. Padding positions remain zero. We initialize $A$ to zero and add the anchor to the phoneme keys and values consumed by every audio block. The relative progress $p_j$ preserves phoneme order without assuming equal phoneme durations.

\subsection{Coupled Speech--Motion Flow Transformer}
\noindent\textbf{Native-rate dual streams.}
The motion and audio latents are projected into two separate Transformer streams. Within each block, each stream first models its own temporal context and reads its modality-adapted text condition. The streams then exchange information bidirectionally through audio-to-motion and motion-to-audio cross-attention, with timestep-conditioned gates regulating the interaction strength throughout denoising.

\noindent\textbf{A common temporal coordinate.}
For the 10-Hz motion latent sequence, token $i$ corresponds to the interval $[i/f_m,(i+1)/f_m)$, where $f_m=10$ Hz. For audio, the patchifier maps each latent index back to a waveform-time interval according to the sampling rate, mel-spectrogram hop, temporal downsampling of the audio VAE, and causal offset, forming a nominal 25-Hz latent clock. Given the different native token rates of speech and motion, we use the midpoint of each interval to construct temporal rotary positional embeddings for within-modality self-attention and bidirectional cross-modal attention. For the same paired motion and audio sample, we also share the flow time $\sigma$, placing the two streams at the same denoising stage. The shared clock, shared noise level, and bidirectional attention together enable the model to learn synchronization from paired data.

\subsection{Joint Flow and Phoneme-Alignment Objectives}
For each modality $r\in\{m,a\}$, let $z_0^r$ be a clean latent and $\epsilon^r\sim\mathcal{N}(0,I)$ independent Gaussian noise. Using one $\sigma\in(0,1)$ for the paired streams, we form the linear probability path
\begin{equation}
z_\sigma^r=(1-\sigma)z_0^r+\sigma\epsilon^r,
\qquad
u^r=\epsilon^r-z_0^r.
\label{eq:flow_path}
\end{equation}
Writing $Z_\sigma=(z_\sigma^m,z_\sigma^a)$, the coupled transformer predicts a modality-specific velocity field $v_\theta^r$ from the joint noisy state and the selected condition $c_\ell$. It is trained with
\begin{equation}
\mathcal{L}_{\mathrm{flow}}=\sum_{r\in\{m,a\}}\lambda_r\,
\mathbb{E}\!\left[\left\|v_\theta^r(Z_\sigma,c_\ell,\sigma)-u^r\right\|_2^2\right].
\label{eq:joint_flow_loss}
\end{equation}
The flow objective learns unimodal fidelity and cross-modal coordination, but it does not explicitly require the intermediate audio sequence to follow transcript order. Let $\mathcal{B}=\{0,\ldots,7\}$ denote the supervised audio blocks and $h_t^{a,b}$ the audio hidden state after block $b$. Each block-specific linear head produces logits over the 183 phoneme symbols plus one dedicated blank, and CTC~\citep{graves2006ctc} marginalizes their monotonic alignments to the target sequence $\phi(s)$:
\begin{equation}
\begin{aligned}
\ell_t^b &= W_{\mathrm{CTC}}^b h_t^{a,b}+b_{\mathrm{CTC}}^b,\\
\mathcal{L}_{\mathrm{CTC}} &=
\frac{1}{|\mathcal{B}|}\sum_{b\in\mathcal{B}}
\operatorname{CTC}\!\left(\operatorname{logsoftmax}(\ell_{1:T}^{b}),\phi(s)\right),\\
\mathcal{L}_{\mathrm{total}} &= \mathcal{L}_{\mathrm{flow}}
+\lambda_{\mathrm{CTC}}\mathcal{L}_{\mathrm{CTC}}.
\end{aligned}
\label{eq:ctc_loss}
\end{equation}
We remove padded phoneme IDs from the targets and derive input lengths from the valid audio-latent mask. The CTC head is used only during training, whereas the learned progress anchor always remains in the phoneme-conditioning path.

\noindent\textbf{Training regimes.}
The model adopts a progressive two-stage training regime. Stage~1 disables the motion stream and cross-modal interaction, and trains the speech and pronunciation pathways on speech-only data using audio flow matching and phoneme-CTC supervision. Stage~2 activates both streams and their bidirectional interaction, optimizing both flow-matching losses and the phoneme-CTC regularizer on the expanded \beatc{} conditioning instances. To preserve the speech branch's generalization ability, we sample 10\% of Stage~2 training batches from the Stage~1 speech-pretraining dataset.

\noindent\textbf{Inference.}
Given a transcript, a voice description, and one L1 or L2 motion instruction, we form $c_\ell$ and encode it with the text encoder to obtain the semantic condition. The transcript is also converted into a phoneme sequence through grapheme-to-phoneme (G2P) conversion and encoded by the frozen phoneme encoder to provide the audio stream's phoneme context. For a target duration, we set the latent lengths at the native motion and audio rates and initialize both streams with independent Gaussian noise. The coupled transformer jointly updates the two latent sequences over a shared noise schedule from $\sigma=1$ to $0$ using 30 Euler steps with classifier-free guidance. Finally, the frozen motion decoder reconstructs 30-Hz SMPL-X motion, while the frozen audio VAE decoder and vocoder reconstruct a 16-kHz waveform matched to the target duration. The phoneme-progress anchor remains active during inference; the auxiliary CTC heads are removed.

\section{Experiments}
\label{sec:experiments}

\subsection{Setup}
\label{sec:setup}

\textbf{Data.} \beatc{} contains 34{,}950 12.9-second segments (approximately 125.2 hours), including 17{,}907 BEAT2 clips and 17{,}043 Seamless Interaction clips. The fixed split contains 27{,}957 training, 3{,}501 validation, and 3{,}492 test segments.

\textbf{Baselines.} CosyVoice~2~\citep{du2024cosyvoice2} followed by MambaTalk~\citep{xu2024mambatalk} forms the cascade baseline. Match-TTSG~\citep{mehta2024match} jointly synthesizes speech and gesture through monotonic phoneme alignment and conditional flow matching. Motion-Omni~\citep{ma2026motionomni} is an end-to-end spoken-dialogue model that jointly outputs speech and full-body motion.

\textbf{Metrics.} We use seven headline metrics. For speech, WER is computed with Whisper large-v3-turbo~\citep{radford2023robust}, while SpkSim is the cosine similarity between WavLM-base-plus-SV~\citep{chen2022wavlm} embeddings of generated and reference speech and serves as a proxy for reference-speaker timbre similarity. For motion, Fr\'{e}chet Gesture Distance (FGD) uses the official EMAGE/PantoMatrix AESKConv features~\citep{liu2024emage}, and foot skating is the percentage of contact frames whose horizontal foot speed exceeds 0.10\,m/s. BC Gap is the absolute mean-BC difference between generated and reference motions, both paired with reference speech. L1 and L2 Global PairAcc use the same frozen CLaM-H text--motion evaluator~\citep{chen2024clam,li2025madaptor}.

\textbf{Implementation details.} \model{} uses eight coupled Transformer blocks for both motion and audio, with width 1{,}024, 32 attention heads, and head dimension 32 in each stream. Frozen modality codecs provide 256-channel motion and 128-channel audio latents. A frozen Gemma-3 encoder~\citep{gemmateam2025gemma3} supplies multi-level text features through cross-attention, while residual connectors adapt them to the two streams. A frozen AudioLDM2 phoneme encoder~\citep{liu2024audioldm2}, the progress anchor, and auxiliary CTC heads at all eight audio blocks provide pronunciation and temporal supervision. Stage~1 trains the audio stream on the Emilia~\citep{he2024emilia} subset for 100{,}000 steps, and Stage~2 jointly trains both streams on \beatc{} for 200{,}000 steps. Both stages use eight A100 GPUs, 8-bit AdamW, a per-GPU batch size of 4, 2{,}000 warm-up steps, and cosine scheduling. The Stage~1 learning rate is $2\!\times\!10^{-4}$; Stage~2 uses $2\!\times\!10^{-4}$ for the motion branch and $1\!\times\!10^{-5}$ for the audio branch. Inference uses 30 Euler steps.

\begin{figure*}[t]
\centering
\includegraphics[width=\textwidth]{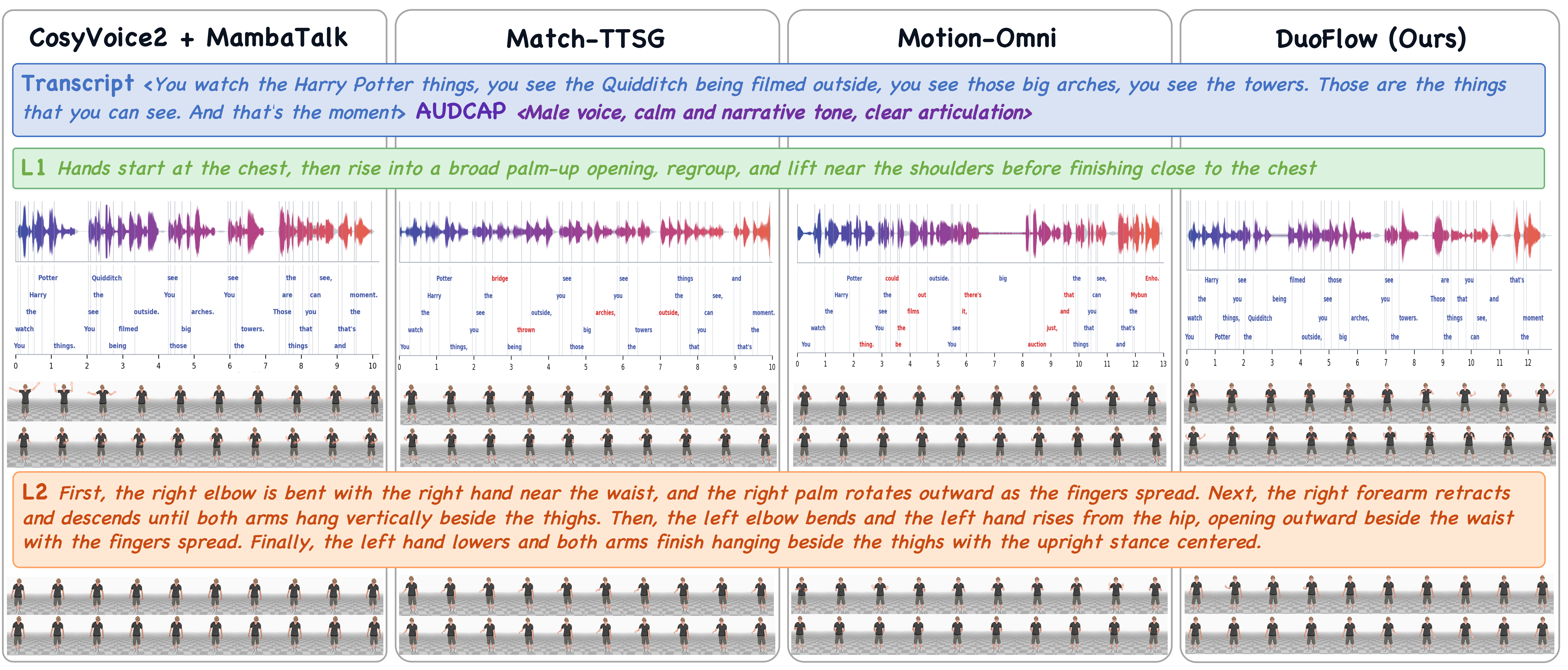}
\caption{\textbf{Qualitative comparison under hierarchical motion control.} The upper block shows generated waveforms, time-aligned transcript words, and sampled motion frames under the same coarse L1 motion instruction. The lower block shows the results after changing the motion description to a fine-grained, temporally ordered L2 instruction. Red words in the time-aligned transcripts mark mismatches with the input text.}
\label{fig:qualitative_original}
\end{figure*}

\begin{figure*}[t]
\centering
\includegraphics[width=\textwidth]{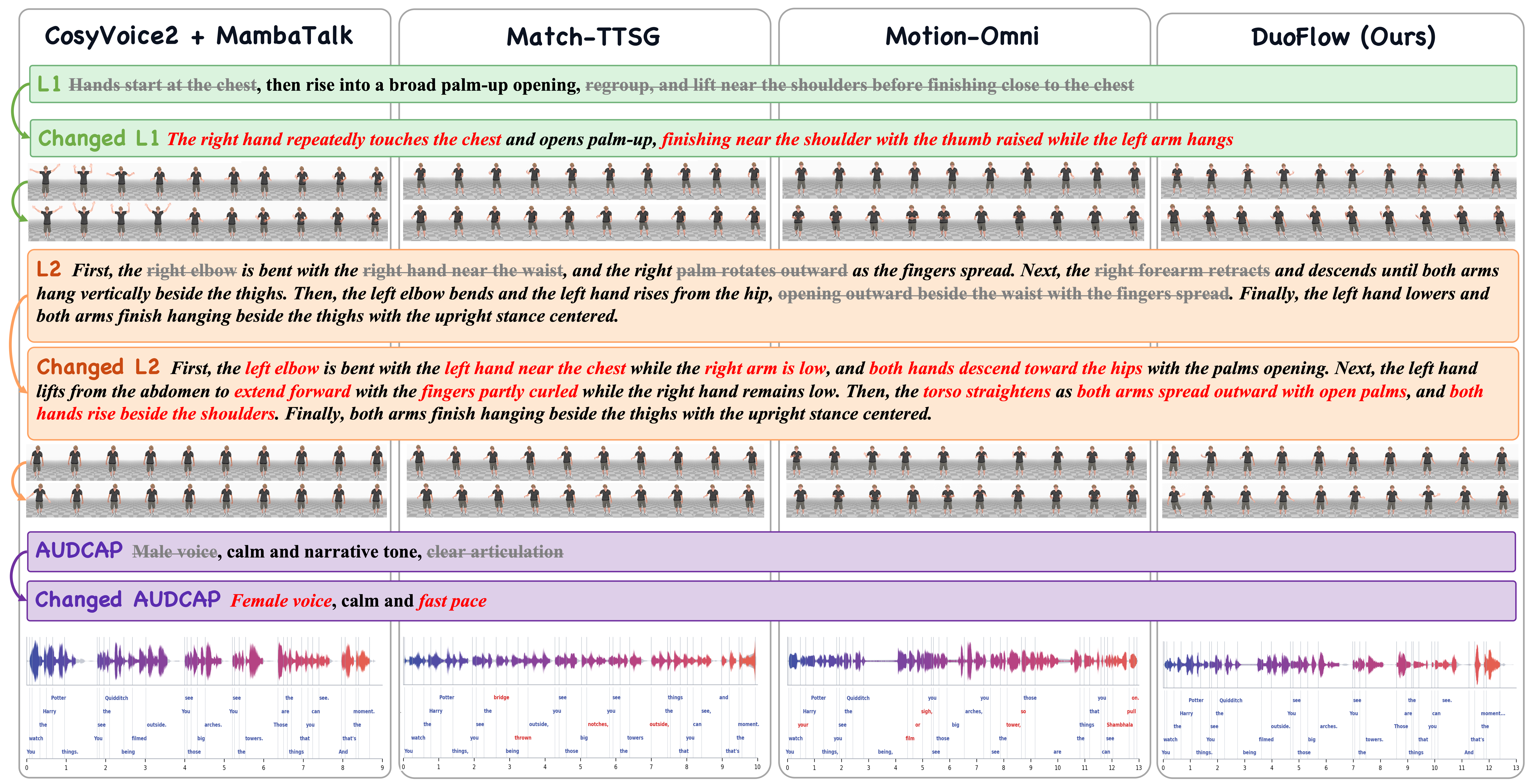}
\caption{\textbf{Single-condition intervention comparison.} In the L1 (green) and L2 (orange) panels, paired motion rows show results before (top) and after (bottom) editing the instruction. Gray strikethroughs and red text mark deleted and added prompt content, respectively. Each intervention keeps the transcript and the other conditions fixed.}
\label{fig:qualitative_interventions}
\end{figure*}

\subsection{Generation Quality, Control, and Efficiency}
\label{sec:structured_control}
We compare speech and motion quality, instruction compatibility, and speech--motion alignment across systems (Table~\ref{tab:main}), and report generation latency in Table~\ref{tab:inference_speed}. Figure~\ref{fig:qualitative_original} complements these metrics with qualitative examples under coarse L1 and fine-grained L2 motion instructions.
\textbf{Speech quality.} Table~\ref{tab:main} shows that \model{} achieves the lowest WER (4.71\%) and highest SpkSim (0.94) among the reported systems. The CosyVoice~2--MambaTalk cascade, built on CosyVoice~2's pretrained speech-synthesis model, maintains good WER and preserves speaker similarity when using a reference audio for timbre. The lightweight Match-TTSG does not generalize its speech capability. Although Motion-Omni, for which no loadable weights are available, uses the same two-stage training strategy as \model{}, an inefficient learning pathway leads to the performance gap, while \model{}'s phoneme pathway provides a stronger representation for the speech branch.

\textbf{Motion quality and instruction compatibility.} \model{} obtains the lowest FGD (0.75) and foot-skating rate (29.61\%), alongside L1/L2 PairAcc of 87.43\%/90.40\%. The structured condition supplies either a global or a detailed motion description to both streams, and the observed scores are consistent with generated motion matching both levels of description. For MambaTalk, which does not accept L1/L2, the scores measure compatibility rather than instruction following. Match-TTSG remains limited by model capacity.

\textbf{Speech--motion alignment.} \model{} combines coupled streams, a shared temporal coordinate, and a joint flow objective, and achieves a BC Gap of 0.14, indicating strong generation consistency. Match-TTSG achieves the lowest BC Gap (0.06), effectively capturing speech--motion rhythm; however, its weaker spatial, physical, and instruction-level motion fidelity shows that beat alignment does not equal natural motion trajectories or good semantic matching.

\begin{table}[t]
\centering
\caption{\textbf{Speech--motion generation quality, alignment, and controllability.} WER, Skate, and PairAcc are percentages. Spk, BC, L1, and L2 abbreviate SpkSim, BC Gap, L1 PairAcc, and L2 PairAcc, respectively. Best and second-best results are shown in bold and underlined, respectively.}
\label{tab:main}
\scriptsize
\setlength{\tabcolsep}{1.35pt}
\renewcommand{\arraystretch}{0.96}
\resizebox{\columnwidth}{!}{%
\begin{tabular}{@{}lrrrrrrr@{}}
\toprule
System & WER$\downarrow$ & Spk$\uparrow$ & FGD$\downarrow$ & Skate$\downarrow$ & BC$\downarrow$ & L1$\uparrow$ & L2$\uparrow$ \\
\midrule
CosyVoice2+Mamba. & \underline{6.12} & \underline{0.80} & 2.61 & \underline{46.78} & 0.23 & 53.42 & 53.63 \\
Match-TTSG & 36.35 & 0.75 & 5.14 & 84.01 & \textbf{0.06} & 56.38 & 60.38 \\
Motion-Omni & 74.98 & 0.56 & \underline{1.32} & 66.57 & 0.34 & \underline{61.37} & \underline{62.68} \\
\midrule
DuoFlow (Ours) & \textbf{4.71} & \textbf{0.94} & \textbf{0.75} & \textbf{29.61} & \underline{0.14} & \textbf{87.43} & \textbf{90.40} \\
\bottomrule
\end{tabular}
}
\end{table}

\textbf{Inference efficiency.} With 30 inference sampling steps, the \model{} dual-stream generation paradigm synthesizes a 12.9-s clip in 1.3298\,s on average (P95 1.5052\,s; RTF 0.1031). Its end-to-end architecture is substantially faster than the reported cascade and Motion-Omni times (Table~\ref{tab:inference_speed}). Joint latent generation incurs considerable overhead.
\begin{table}[t]
\centering
\caption{\textbf{Inference speed.} $T_{\mathrm{resp}}$ is mean time to generate one 12.9-s speech--motion clip at batch size one; RTF is $T_{\mathrm{resp}}/12.9$; Slowdown is relative to the fastest reported system.}
\label{tab:inference_speed}
\small
\setlength{\tabcolsep}{4pt}
\resizebox{\columnwidth}{!}{%
\begin{tabular}{lccc}
\toprule
Model & $T_{\mathrm{resp}}$ (s)$\downarrow$ & RTF$\downarrow$ & Slowdown$\downarrow$ \\
\midrule
CosyVoice2 + MambaTalk & 2.82 & 0.22 & 8.07$\times$ \\
Match-TTSG & \textbf{0.35} & \textbf{0.03} & \textbf{1.00$\times$} \\
Motion-Omni & 5.03 & 0.39 & 14.39$\times$ \\
DuoFlow (ours) & \underline{1.33} & \underline{0.10} & \underline{3.80$\times$} \\
\bottomrule
\end{tabular}%
}
\end{table}

\textbf{Single-condition interventions.} Figure~\ref{fig:qualitative_interventions} compares the systems under separate edits to L1, L2, or AUDCAP while keeping the transcript and the other conditions fixed. \model{} is the only generative model that responds well to these control changes.

\subsection{Ablation Studies}
\label{sec:ablations}
\textbf{Phoneme conditioning.} Table~\ref{tab:stage1_ablation} isolates changes to the Stage~1 phoneme path. Adding the phoneme branch reduces WER from 9.45\% to 6.02\%, suggesting that an explicit pronunciation representation complements the shared semantic text condition. Adding the progress anchor together with multi-depth CTC further reduces WER to 5.29\%. The anchor supplies relative phoneme order, while CTC encourages a monotonic transcript-compatible path in intermediate audio states.
\begin{table}[t]
\centering
\caption{\textbf{Stage-1 audio-path ablation.} WER is evaluated on \beatc{} after Stage~1 training.}
\label{tab:stage1_ablation}
\small
\setlength{\tabcolsep}{2.6pt}
\resizebox{\columnwidth}{!}{%
\begin{tabular}{lcccc}
\toprule
Configuration & Phoneme & Anchor & CTC & WER$\downarrow$ \\
\midrule
Text condition only & -- & -- & -- & 9.45 \\
$+$ phoneme branch & \checkmark & -- & -- & \underline{6.02} \\
$+$ CTC and anchor & \checkmark & \checkmark & \checkmark & \textbf{5.29} \\
\bottomrule
\end{tabular}%
}
\end{table}

\textbf{Training schedule.} Compared with single-stage joint training, audio pretraining followed by joint training reduces WER from 38.76\% to 4.71\% and improves SpkSim from 0.82 to 0.94 (Table~\ref{tab:two_stage_training_ablation}). The same schedule improves FGD, BC Gap, and both PairAcc scores. This result indicates that learning an effective speech prior before modality coupling is necessary and is consistent with preserving motion quality and description compatibility during joint training.
\begin{table}[t]
\centering
\caption{\textbf{Training-schedule ablation.} Single-stage joint training versus cascaded training, with audio pretraining followed by joint training.}
\label{tab:two_stage_training_ablation}
\small
\setlength{\tabcolsep}{2.2pt}
\resizebox{\columnwidth}{!}{%
\begin{tabular}{lcccccc}
\toprule
Strategy & WER$\downarrow$ & SpkSim$\uparrow$ & FGD$\downarrow$ & BC Gap$\downarrow$ & L1 PairAcc$\uparrow$ & L2 PairAcc$\uparrow$ \\
\midrule
Single-stage joint & \underline{38.76} & \underline{0.82} & \underline{0.98} & \underline{0.18} & \underline{80.24} & \underline{83.17} \\
Cascaded training & \textbf{4.71} & \textbf{0.94} & \textbf{0.75} & \textbf{0.14} & \textbf{87.43} & \textbf{90.40} \\
\bottomrule
\end{tabular}%
}
\end{table}

\section{Conclusion}
\label{sec:conclusion}

We presented \beatc{}, a four-channel benchmark that aligns a transcript and vocal-style description with coarse- and fine-grained motion intents, together with \model{}, a dual-stream latent flow framework for controllable joint speech--motion generation. The framework coordinates native-rate speech and motion latents through a shared noise level, a common second-based temporal coordinate, modality-specific residual connectors, and layer-wise bidirectional interaction. It strengthens speech temporal awareness through a monotonic phoneme-alignment pathway that combines a learned normalized progress anchor with training-only CTC supervision. The results show that an end-to-end model can produce intelligible speech, realistic and synchronized full-body motion, consistency with both levels of motion instruction, and controllable vocal delivery. This work establishes a unified interface for multimodal motion and speech control, while improving physical contact fidelity and extending the framework to streaming generation remain future directions.

\section*{LLM Usage Statement}
ChatGPT and OpenAI Codex were used to assist with language editing, manuscript revision, code development, and debugging.
GLM-5.3-Flash was used to generate initial L1/L2 motion descriptions from rendered motion videos, and
Qwen3-Omni-30B-A3B-Instruct was used to generate initial AUDCAP descriptions from speech waveforms, as described in
Section~3.2. The authors remain responsible for all submitted text, code, annotations, figures, and scientific claims.

{
  \small
  \bibliographystyle{ieeenat_fullname}
  \bibliography{references}

@misc{ma2026motionomni,
  title         = {Motion-Omni: End-to-End Joint Speech and Full-Body Motion for Spoken Dialogue},
  author        = {Ma, Chengqian and Tao, Wei and Zhang, Haoyu and Guo, Yiwen},
  year          = {2026},
  eprint        = {2609.04250},
  archivePrefix = {arXiv},
  primaryClass  = {cs.CL},
  url           = {https://arxiv.org/abs/2609.04250},
  note          = {arXiv:2609.04250}
}

@inproceedings{mehta2024match,
  title     = {Unified Speech and Gesture Synthesis Using Flow Matching (Match-TTSG)},
  author    = {Mehta, Shivam and Tu, Ruibo and Alexanderson, Simon and Beskow, Jonas and Sz{\'e}kely, {\'E}va and Henter, Gustav Eje},
  booktitle = {IEEE International Conference on Acoustics, Speech and Signal Processing (ICASSP)},
  year      = {2024},
  url       = {https://arxiv.org/abs/2310.05181},
}

@inproceedings{guo2024fasttalker,
  title     = {FastTalker: Jointly Generating Speech and Conversational Gestures from Text},
  author    = {Guo, Zixin and Zhang, Jian},
  booktitle = {European Conference on Computer Vision (ECCV) Workshops},
  year      = {2024},
  url       = {https://arxiv.org/abs/2409.16404},
}

@inproceedings{guichoux2026gelina,
  title     = {Gelina: Unified Speech and Gesture Synthesis via Interleaved Token Prediction},
  author    = {Guichoux, T{\'e}o and Lemerle, Th{\'e}odor and Mehta, Shivam and Beskow, Jonas and others},
  booktitle = {IEEE International Conference on Acoustics, Speech and Signal Processing (ICASSP)},
  pages     = {16122--16126},
  year      = {2026},
  doi       = {10.1109/ICASSP55912.2026.11464562},
  url       = {https://arxiv.org/abs/2510.12834},
}

@misc{alexanderson2021spontaneous,
  title         = {Generating Coherent Spontaneous Speech and Gesture from Text},
  author        = {Alexanderson, Simon and Sz{\'e}kely, {\'E}va and Henter, Gustav Eje and Kucherenko, Taras and Beskow, Jonas},
  year          = {2021},
  eprint        = {2101.05684},
  archivePrefix = {arXiv},
  url           = {https://arxiv.org/abs/2101.05684},
  note          = {arXiv:2101.05684}
}

@misc{wang2021integrated,
  title         = {Integrated Speech and Gesture Synthesis},
  author        = {Wang, Siyang and Alexanderson, Simon and Gustafson, Joakim and Beskow, Jonas and Henter, Gustav Eje and Sz{\'e}kely, {\'E}va},
  year          = {2021},
  eprint        = {2108.11436},
  archivePrefix = {arXiv},
  url           = {https://arxiv.org/abs/2108.11436},
  note          = {arXiv:2108.11436}
}

@misc{chen2024syntalker,
  title         = {Enabling Synergistic Full-Body Control in Prompt-Based Co-Speech Motion Generation},
  author        = {Chen, Bohong and Li, Yumeng and Ding, Yao-Xiang and Shao, Tianjia and Zhou, Kun},
  year          = {2024},
  eprint        = {2410.00464},
  archivePrefix = {arXiv},
  url           = {https://arxiv.org/abs/2410.00464},
  note          = {arXiv:2410.00464}
}

@misc{pinyoanuntapong2026interactgesture,
  title         = {{InteractGesture}: Progressive Chunk Guidance for Continuous Streaming Co-Speech Gesture Control},
  author        = {Pinyoanuntapong, Ekkasit and Deogade, Ajinkya and Streli, Paul and Zhang, Wenjing and Materzynska, Joanna and Wang, Pu and Ferrari, Vittorio and Shen, Jie},
  year          = {2026},
  eprint        = {2608.25734},
  archivePrefix = {arXiv},
  url           = {https://arxiv.org/abs/2608.25734},
  note          = {arXiv:2608.25734}
}

@misc{lim2026faceplex,
  title         = {{FacePlex}: Full-Duplex Joint Speech-Facial Motion Generation for Conversational Avatars},
  author        = {Lim, Habin and Lee, Jae-Ho and Lew, Hah Min and Kang, Ji-Su and Park, Gyeong-Moon},
  year          = {2026},
  eprint        = {2606.30145v1},
  archivePrefix = {arXiv},
  url           = {https://arxiv.org/abs/2606.30145v1},
  note          = {arXiv:2606.30145v1}
}

@misc{nagy2026genea,
  title         = {The {GENEA} Challenge 2026: A Large-Scale Disentangled Evaluation of Speech-Driven Gesture Generation on the Seamless Interaction Dataset},
  author        = {Nagy, Rajmund and Arellano Garc{\'i}a, Silvia and Voss, Hendric and Tsakov, Mihail and Kucherenko, Taras and Yoon, Youngwoo and Henter, Gustav Eje},
  year          = {2026},
  eprint        = {2608.10839},
  archivePrefix = {arXiv},
  url           = {https://arxiv.org/abs/2608.10839},
  note          = {arXiv:2608.10839}
}

@misc{du2024cosyvoice2,
  title         = {CosyVoice 2: Scalable Streaming Speech Synthesis with Large Language Models},
  author        = {Du, Zhihao and Wang, Yuxuan and Chen, Qian and others},
  year          = {2024},
  eprint        = {2412.10117},
  archivePrefix = {arXiv},
  primaryClass  = {cs.CL},
  url           = {https://arxiv.org/abs/2412.10117},
  note          = {arXiv:2412.10117}
}

@inproceedings{xu2024mambatalk,
  title     = {MambaTalk: Efficient Holistic Gesture Synthesis with Selective State Space Models},
  author    = {Xu, Zunnan and Lin, Yukang and Han, Haonan and Yang, Sicheng and others},
  booktitle = {Advances in Neural Information Processing Systems (NeurIPS)},
  year      = {2024},
  url       = {https://arxiv.org/abs/2403.09471},
}

@inproceedings{yang2025unimumo,
  title     = {UniMuMo: Unified Text, Music and Motion Generation},
  author    = {Yang, Han and Su, Kun and Zhang, Yutong and others},
  booktitle = {Proceedings of the AAAI Conference on Artificial Intelligence},
  year      = {2025},
  url       = {https://arxiv.org/abs/2410.04534},
}

@inproceedings{you2024momu,
  title     = {MoMu-Diffusion: On Learning Long-Term Motion-Music Synchronization and Correspondence},
  author    = {You, Fuming and Fang, Minghui and Tang, Li and Huang, Rongjie and others},
  booktitle = {Advances in Neural Information Processing Systems (NeurIPS)},
  year      = {2024},
  url       = {https://arxiv.org/abs/2411.01805},
}

@inproceedings{kwon2026jamflow,
  title     = {JAM-Flow: Joint Audio-Motion Synthesis with Flow Matching},
  author    = {Kwon, Mingi and Shin, Joonghyuk and Jeong, Jaeseok and Park, Jaesik and Uh, Youngjung},
  booktitle = {European Conference on Computer Vision (ECCV)},
  pages     = {353--371},
  volume    = {17020},
  series    = {Lecture Notes in Computer Science},
  publisher = {Springer},
  year      = {2026},
  doi       = {10.1007/978-3-032-37553-7_20},
  url       = {https://doi.org/10.1007/978-3-032-37553-7_20},
}

@inproceedings{jang2024faces,
  title     = {Faces that Speak: Jointly Synthesising Talking Face and Speech from Text},
  author    = {Jang, Youngjoon and Kim, Ji-Hoon and Ahn, Junseok and others},
  booktitle = {IEEE/CVF Conference on Computer Vision and Pattern Recognition (CVPR)},
  year      = {2024},
  url       = {https://arxiv.org/abs/2405.10272},
}

@misc{cheng2026unison,
  title         = {Unison: Harmonizing Motion, Speech, and Sound for Human-Centric Audio-Video Generation},
  author        = {Cheng, Shihao and Zhang, Jiaxu and Song, Quanyue and others},
  year          = {2026},
  eprint        = {2605.08729},
  archivePrefix = {arXiv},
  primaryClass  = {cs.CV},
  url           = {https://arxiv.org/abs/2605.08729},
  note          = {arXiv:2605.08729}
}

@misc{wang2025mgmomni,
  title         = {MGM-Omni: Scaling Omni LLMs to Personalized Long-Horizon Speech},
  author        = {Wang, Chengyao and Zhong, Zhisheng and Peng, Bohao and others},
  year          = {2025},
  eprint        = {2509.25131},
  archivePrefix = {arXiv},
  primaryClass  = {cs.CL},
  url           = {https://arxiv.org/abs/2509.25131},
  note          = {arXiv:2509.25131}
}

@inproceedings{chen2025rapverse,
  title     = {RapVerse: Coherent Vocals and Whole-Body Motions Generations from Text},
  author    = {Chen, Jiaben and Yan, Xin and Chen, Yihang and others},
  booktitle = {IEEE/CVF International Conference on Computer Vision (ICCV)},
  year      = {2025},
  url       = {https://arxiv.org/abs/2405.20336},
}

@inproceedings{liu2022beat,
  title     = {{BEAT}: A Large-Scale Semantic and Emotional Multi-Modal Dataset for Conversational Gestures Synthesis},
  author    = {Liu, Haiyang and Zhu, Zihao and Iwamoto, Naoya and others},
  booktitle = {European Conference on Computer Vision (ECCV)},
  year      = {2022},
  url       = {https://arxiv.org/abs/2203.05297},
}

@misc{agrawal2025seamless,
  title         = {Seamless Interaction: Dyadic Audiovisual Motion Modeling and Large-Scale Dataset},
  author        = {Agrawal, Vasu and Akinyemi, Akinniyi and Alvero, Kathryn and Behrooz, Morteza and Buffalini, Julia and Carlucci, Fabio Maria and others},
  year          = {2025},
  eprint        = {2506.22554},
  archivePrefix = {arXiv},
  primaryClass  = {cs.CV},
  url           = {https://arxiv.org/abs/2506.22554},
  note          = {arXiv:2506.22554}
}

@inproceedings{liu2024emage,
  title     = {{EMAGE}: Towards Unified Holistic Co-Speech Gesture Generation via Expressive Masked Audio Gesture Modeling},
  author    = {Liu, Haiyang and Zhu, Zihao and Becherini, Giorgio and others},
  booktitle = {IEEE/CVF Conference on Computer Vision and Pattern Recognition (CVPR)},
  year      = {2024},
  url       = {https://arxiv.org/abs/2401.00374},
}

@misc{kucherenko2023genea,
  title         = {The {GENEA} Challenge 2023: A Large Scale Evaluation of Gesture Generation Models in Monadic and Dyadic Settings},
  author        = {Kucherenko, Taras and Nagy, Rajmund and Yoon, Youngwoo and Woo, Jieyeon and Nikolov, Teodor and Tsakov, Mihail and Henter, Gustav Eje},
  year          = {2023},
  eprint        = {2308.12646},
  archivePrefix = {arXiv},
  url           = {https://arxiv.org/abs/2308.12646},
  note          = {arXiv:2308.12646}
}

@inproceedings{ginosar2019learning,
  title     = {Learning Individual Styles of Conversational Gesture},
  author    = {Ginosar, Shiry and Bar, Amir and Kohavi, Gefen and others},
  booktitle = {IEEE/CVF Conference on Computer Vision and Pattern Recognition (CVPR)},
  year      = {2019},
  url       = {https://arxiv.org/abs/1906.04160},
}

@inproceedings{ferstl2018recurrent,
  title     = {Investigating the Use of Recurrent Motion Modelling for Speech Gesture Generation},
  author    = {Ferstl, Ylva and McDonnell, Rachel},
  booktitle = {Proceedings of the 18th International Conference on Intelligent Virtual Agents},
  pages     = {93--98},
  year      = {2018},
  doi       = {10.1145/3267851.3267898}
}

@inproceedings{lee2019talking,
  title     = {Talking With Hands 16.2M: A Large-Scale Dataset of Synchronized Body-Finger Motion and Audio for Conversational Motion Analysis and Synthesis},
  author    = {Lee, Gilwoo and Deng, Zhiwei and Ma, Shugao and Shiratori, Takaaki and Srinivasa, Siddhartha S. and Sheikh, Yaser},
  booktitle = {IEEE/CVF International Conference on Computer Vision (ICCV)},
  pages     = {763--772},
  year      = {2019}
}

@inproceedings{ahuja2020style,
  title     = {Style Transfer for Co-Speech Gesture Animation: A Multi-Speaker Conditional-Mixture Approach},
  author    = {Ahuja, Chaitanya and Lee, Dong Won and Nakano, Yukiko I. and Morency, Louis-Philippe},
  booktitle = {European Conference on Computer Vision (ECCV)},
  pages     = {248--265},
  year      = {2020},
  url       = {https://arxiv.org/abs/2007.12553}
}

@article{yoon2020speech,
  title   = {Speech Gesture Generation from the Trimodal Context of Text, Audio, and Speaker Identity},
  author  = {Yoon, Youngwoo and Cha, Bok and Lee, Joo-Haeng and others},
  journal = {ACM Transactions on Graphics (SIGGRAPH Asia 2020)},
  volume  = {39},
  number  = {6},
  year    = {2020},
  doi     = {10.1145/3414685.3417838},
  url     = {https://arxiv.org/abs/2009.02119},
}

@inproceedings{liu2025gesturelsm,
  title     = {{GestureLSM}: Latent Shortcut based Co-Speech Gesture Generation with Spatial-Temporal Modeling},
  author    = {Liu, Pinxin and Song, Luchuan and Huang, Junhua and Liu, Haiyang and others},
  booktitle = {IEEE/CVF International Conference on Computer Vision (ICCV)},
  year      = {2025},
  url       = {https://arxiv.org/abs/2501.18898},
}

@misc{hacohen2026ltx2,
  title         = {{LTX-2}: Efficient Joint Audio-Visual Foundation Model},
  author        = {HaCohen, Yoav and Brazowski, Benny and Chiprut, Nisan and others},
  year          = {2026},
  eprint        = {2601.03233},
  archivePrefix = {arXiv},
  primaryClass  = {cs.CV},
  url           = {https://arxiv.org/abs/2601.03233},
  note          = {arXiv:2601.03233}
}

@inproceedings{kim2021vits,
  title     = {Conditional Variational Autoencoder with Adversarial Learning for End-to-End Text-to-Speech},
  author    = {Kim, Jaehyeon and Kong, Jungil and Son, Juhee},
  booktitle = {International Conference on Machine Learning (ICML)},
  year      = {2021},
  url       = {https://arxiv.org/abs/2106.06103}
}

@misc{gemmateam2025gemma3,
  title         = {{Gemma 3 Technical Report}},
  author        = {{Gemma Team} and Kamath, Aishwarya and Ferret, Johan and Pathak, Shreya and Vieillard, Nino and Merhej, Ramona and others},
  year          = {2025},
  eprint        = {2503.19786},
  archivePrefix = {arXiv},
  primaryClass  = {cs.CL},
  url           = {https://arxiv.org/abs/2503.19786},
  note          = {arXiv:2503.19786}
}

@inproceedings{chen2024diffsheg,
  title     = {{DiffSHEG}: A Diffusion-Based Approach for Real-Time Speech-driven Holistic 3D Expression and Gesture Generation},
  author    = {Chen, Junming and Liu, Yunfei and Wang, Jianan and others},
  booktitle = {IEEE/CVF Conference on Computer Vision and Pattern Recognition (CVPR)},
  year      = {2024},
  url       = {https://arxiv.org/abs/2401.04747},
}

@inproceedings{zhang2026vibes,
  title     = {{ViBES}: A Conversational Agent with Behaviorally-Intelligent 3D Virtual Body},
  author    = {Zhang, Juze and Chen, Changan and Chen, Xin and others},
  booktitle = {IEEE/CVF Conference on Computer Vision and Pattern Recognition (CVPR)},
  year      = {2026},
  url       = {https://arxiv.org/abs/2512.14234},
}

@misc{nagano2026dyaplex,
  title         = {{DyaPlex}: Full-Duplex Speech-Motion Model for Dyadic Interaction},
  author        = {Nagano, Koki and Liu, Hongyu and Park, Seonwook and others},
  year          = {2026},
  eprint        = {2606.03874},
  archivePrefix = {arXiv},
  primaryClass  = {cs.CV},
  url           = {https://arxiv.org/abs/2606.03874},
  note          = {arXiv:2606.03874}
}

@inproceedings{deng2026umind,
  title     = {{U-Mind}: A Unified Framework for Real-Time Multimodal Interaction with Audiovisual Generation},
  author    = {Deng, Xiang and Gao, Feng and Zhang, Yong and others},
  booktitle = {IEEE/CVF Conference on Computer Vision and Pattern Recognition (CVPR)},
  year      = {2026},
  url       = {https://arxiv.org/abs/2602.23739},
}

@misc{zhang2026exomni,
  title         = {{Ex-Omni}: Enabling 3D Facial Animation Generation for Omni-modal Large Language Models},
  author        = {Zhang, Haoyu and Li, Zhipeng and Guo, Yiwen and Yu, Tianshu},
  year          = {2026},
  eprint        = {2602.07106},
  archivePrefix = {arXiv},
  primaryClass  = {cs.CV},
  url           = {https://arxiv.org/abs/2602.07106},
  note          = {arXiv:2602.07106}
}

@misc{jiang2024solami,
  title         = {{SOLAMI}: Social Vision-Language-Action Modeling for Immersive Interaction with 3D Autonomous Characters},
  author        = {Jiang, Jianping and Xiao, Weiye and Lin, Zhengyu and others},
  year          = {2024},
  eprint        = {2412.00174},
  archivePrefix = {arXiv},
  primaryClass  = {cs.CV},
  url           = {https://arxiv.org/abs/2412.00174},
  note          = {arXiv:2412.00174}
}

@misc{zhu2026omniliveavatar,
  title         = {{Omni-LiveAvatar}: Minute-Level Real-Time Streaming Joint Audio-Video Avatar Generation},
  author        = {Zhu, Lunjie and Ge, Xingtong and Lin, Fangyu and others},
  year          = {2026},
  eprint        = {2608.13602},
  archivePrefix = {arXiv},
  primaryClass  = {cs.CV},
  url           = {https://arxiv.org/abs/2608.13602},
  note          = {arXiv:2608.13602}
}

@inproceedings{ji2026discoforcing,
  title     = {{DiscoForcing}: A Unified Framework for Real-Time Audio-Driven Character Control with Diffusion Forcing},
  author    = {Ji, Kaiyang and Qian, Bingsheng and Wu, Binghuan and others},
  booktitle = {International Conference on Machine Learning (ICML)},
  year      = {2026},
  url       = {https://arxiv.org/abs/2605.28491},
}

@misc{zhao2026personagest,
  title         = {{PersonaGest}: Personalized Co-Speech Gesture Generation with Semantic-Guided Hierarchical Motion Representation},
  author        = {Zhao, Junchuan and Liang, Qifan and Wang, Ye},
  year          = {2026},
  eprint        = {2605.07252},
  archivePrefix = {arXiv},
  primaryClass  = {cs.CV},
  url           = {https://arxiv.org/abs/2605.07252},
  note          = {arXiv:2605.07252}
}

@misc{fang2025coordspeaker,
  title         = {{CoordSpeaker}: Exploiting Gesture Captioning for Coordinated Caption-Empowered Co-Speech Gesture Generation},
  author        = {Fang, Fengyi and Yang, Sicheng and Yang, Wenming},
  year          = {2025},
  eprint        = {2511.22863},
  archivePrefix = {arXiv},
  primaryClass  = {cs.CV},
  url           = {https://arxiv.org/abs/2511.22863},
  note          = {arXiv:2511.22863}
}

@misc{liu2026semconflow,
  title         = {{SemConFlow}: Semantic Grounding of Holistic Co-Speech Gesture Generation with Contrastive Flow-Matching},
  author        = {Liu, Lanmiao and Ghaleb, Esam and {\"O}zy{\"u}rek, Asl{\i} and others},
  year          = {2026},
  eprint        = {2603.26553},
  archivePrefix = {arXiv},
  primaryClass  = {cs.CV},
  url           = {https://arxiv.org/abs/2603.26553},
  note          = {arXiv:2603.26553}
}

@misc{rosenthal2026signature,
  title         = {{SiGnature}: Explicit Motion Diffusion for Stylized Semantic Gesture},
  author        = {Rosenthal, Adi and Koren, Tomer and Shaked, Nadav and others},
  year          = {2026},
  eprint        = {2606.15889},
  archivePrefix = {arXiv},
  primaryClass  = {cs.CV},
  url           = {https://arxiv.org/abs/2606.15889},
  note          = {arXiv:2606.15889}
}

@article{nyatsanga2023review,
  title   = {A Comprehensive Review of Data-Driven Co-Speech Gesture Generation},
  author  = {Nyatsanga, Simbarashe and Kucherenko, Taras and Ahuja, Chaitanya and others},
  journal = {Computer Graphics Forum (Eurographics 2023)},
  volume  = {42},
  number  = {2},
  year    = {2023},
  doi     = {10.1111/cgf.14776},
  url     = {https://arxiv.org/abs/2301.05339},
}

@inproceedings{xiao2025motionstreamer,
  title     = {MotionStreamer: Streaming Motion Generation via Diffusion-Based Autoregressive Model in Causal Latent Space},
  author    = {Xiao, Lixing and Lu, Shunlin and Pi, Huaijin and Fan, Ke and Pan, Liang and Zhou, Yueer and Feng, Ziyong and Zhou, Xiaowei and Peng, Sida and Wang, Jingbo},
  booktitle = {Proceedings of the IEEE/CVF International Conference on Computer Vision (ICCV)},
  pages     = {10086--10096},
  year      = {2025},
  url       = {https://github.com/zju3dv/MotionStreamer}
}

@inproceedings{radford2023robust,
  title     = {Robust Speech Recognition via Large-Scale Weak Supervision},
  author    = {Radford, Alec and Kim, Jong Wook and Xu, Tao and Brockman, Greg and McLeavey, Christine and Sutskever, Ilya},
  booktitle = {Proceedings of the 40th International Conference on Machine Learning},
  pages     = {28492--28518},
  volume    = {202},
  series    = {Proceedings of Machine Learning Research},
  publisher = {PMLR},
  year      = {2023},
  url       = {https://proceedings.mlr.press/v202/radford23a.html}
}

@article{chen2022wavlm,
  title   = {{WavLM}: Large-Scale Self-Supervised Pre-Training for Full Stack Speech Processing},
  author  = {Chen, Sanyuan and Wang, Chengyi and Chen, Zhengyang and others},
  journal = {IEEE Journal of Selected Topics in Signal Processing},
  volume  = {16},
  number  = {6},
  pages   = {1505--1518},
  year    = {2022},
  doi     = {10.1109/JSTSP.2022.3188113}
}

@inproceedings{yi2023talkshow,
  title     = {Generating Holistic 3D Human Motion from Speech},
  author    = {Yi, Hongwei and Liang, Hualin and Liu, Yifei and Cao, Qiong and Wen, Yandong and Bolkart, Timo and Tao, Dacheng and Black, Michael J.},
  booktitle = {Proceedings of the IEEE/CVF Conference on Computer Vision and Pattern Recognition},
  pages     = {469--480},
  year      = {2023},
  doi       = {10.1109/CVPR52729.2023.00053},
  url       = {https://doi.org/10.1109/CVPR52729.2023.00053},
}

@misc{gjaci2026sicage,
  title         = {{SICAGE}: Speaker-Independent Culture-Aware Gesture Generation Using {TED4C-L} Dataset},
  author        = {Gjaci, Ariel and Sgorbissa, Antonio and Murino, Vittorio},
  year          = {2026},
  eprint        = {2606.30001},
  archivePrefix = {arXiv},
  primaryClass  = {cs.CV},
  url           = {https://arxiv.org/abs/2606.30001},
  note          = {arXiv:2606.30001}
}

@inproceedings{graves2006ctc,
  author    = {Alex Graves and Santiago Fern{\'a}ndez and Faustino Gomez and J{\"u}rgen Schmidhuber},
  title     = {Connectionist Temporal Classification: Labelling Unsegmented Sequence Data with Recurrent Neural Networks},
  booktitle = {Proceedings of the 23rd International Conference on Machine Learning},
  pages     = {369--376},
  year      = {2006},
  doi       = {10.1145/1143844.1143891}
}

@inproceedings{chen2024clam,
  title     = {{CLaM}: An Open-Source Library for Performance Evaluation of Text-Driven Human Motion Generation},
  author    = {Chen, Xiaodong and He, Kunlang and Liu, Wu and Liu, Xinchen and Zha, Zheng-Jun and Mei, Tao},
  booktitle = {Proceedings of the 32nd ACM International Conference on Multimedia},
  year      = {2024}
}

@inproceedings{li2025madaptor,
  title     = {{M-Adaptor}: Text-Driven Whole-Body Human Motion Generation},
  author    = {Li, Alicia and Chen, Xiaodong and Liang, Bohao and Bao, Qian and Liu, Wu},
  booktitle = {Proceedings of the IEEE/CVF Conference on Computer Vision and Pattern Recognition (CVPR) Workshops},
  pages     = {2629--2638},
  year      = {2025},
  url       = {https://openaccess.thecvf.com/content/CVPR2025W/PVUW/html/Li_M-Adaptor_Text-driven_Whole-body_Human_Motion_Generation_CVPRW_2025_paper.html}
}

@article{liu2024audioldm2,
  title   = {{AudioLDM 2}: Learning Holistic Audio Generation With Self-Supervised Pretraining},
  author  = {Liu, Haohe and Yuan, Yi and Liu, Xubo and Mei, Xinhao and Kong, Qiuqiang and Tian, Qiao and Wang, Yuping and Wang, Wenwu and Wang, Yuxuan and Plumbley, Mark D.},
  journal = {IEEE/ACM Transactions on Audio, Speech, and Language Processing},
  volume  = {32},
  pages   = {2871--2883},
  year    = {2024},
  doi     = {10.1109/TASLP.2024.3399607},
  url     = {https://doi.org/10.1109/TASLP.2024.3399607}
}

@inproceedings{he2024emilia,
  title     = {{Emilia}: An Extensive, Multilingual, and Diverse Speech Dataset for Large-Scale Speech Generation},
  author    = {He, Haorui and Shang, Zengqiang and Wang, Chaoren and Li, Xuyuan and Gu, Yicheng and Hua, Hua and Liu, Liwei and Yang, Chen and Li, Jiaqi and Shi, Peiyang and Wang, Yuancheng and Chen, Kai and Zhang, Pengyuan and Wu, Zhizheng},
  booktitle = {IEEE Spoken Language Technology Workshop (SLT)},
  pages     = {885--890},
  year      = {2024},
  doi       = {10.1109/SLT61566.2024.10832365},
  url       = {https://arxiv.org/abs/2407.05361}
}
}

\end{document}